\documentclass[10pt,twocolumn,letterpaper]{article}
\usepackage{cvpr}
\usepackage{times}
\usepackage{epsfig}
\usepackage{graphicx}
\usepackage{amsmath}
\usepackage{amssymb}
 
\usepackage[square,numbers,sort&compress]{natbib}
\usepackage[font=small,skip=0pt]{caption}

\usepackage{algorithm}
\usepackage{algorithmic}
\usepackage[pagebackref=false,breaklinks=true,letterpaper=true,colorlinks,bookmarks=false]{hyperref}

\newcommand{\comments}[1]{}
   
\cvprfinalcopy % *** Uncomment this line for the final submission

\def\ackname{Acknowledgements}
\def\acknowledgement{\par\addvspace{17pt}\small\rmfamily
\trivlist\if!\ackname!\item[]\else
\item[\hskip\labelsep
{\bfseries\ackname}]\fi}

\newenvironment{acknowledgements}{\begin{acknowledgement}}
{\end{acknowledgement}}
\def\cvprPaperID{***} % *** Enter the CVPR Paper ID here

\ifcvprfinal\fi
\begin{document}
 
%%%%%%%%% TITLE
\title{Heterogeneous Multi-task Learning for Human Pose Estimation with Deep Convolutional Neural Network}
 
\author{Sijin Li\\
{\small Dept. of Computer Science}\\
{\small City University of Hong Kong}\\
{\tt\small sijin.li@my.cityu.edu.hk}
%sijin.li@my.cityu.edu.hk
\and 
Zhi-Qiang Liu\\
{\small School of Creative Media}\\
{\small City University of Hong Kong}\\
{\tt\small SMZLIU@cityu.edu.hk}
\and
Antoni B. Chan\\
{\small Dept. of Computer Science}\\
{\small City University of Hong Kong}\\
{\tt\small abchan@cityu.edu.hk}
}

\maketitle
%\thispagestyle{empty}
 
%%%%%%%%% ABSTRACT
\begin{abstract} 
\vspace{-0.1in}
We propose an heterogeneous multi-task learning framework for human pose estimation from monocular image with deep convolutional neural network. In particular, we simultaneously learn a pose-joint regressor and a sliding-window body-part detector in a deep network architecture.  We show that including the body-part detection task helps to regularize the network, directing it to converge to a good solution. We report competitive and state-of-art results on several data sets. 
We also empirically show that the learned neurons in the middle layer of our network are tuned to localized body parts. %shape selective.   
\end{abstract}

%%%%%%%%% BODY TEXT

\vspace{-0.1in}
\section{Introduction} 
Human pose estimation is a popular research topic in computer vision for its wide potential in many applications, such as video games, gesture control, action understanding, pose retrieval. Human pose estimation from depth images is much more mature than estimation from 2D image. Some algorithms~\cite{Shotton_2011} based on depth maps have already been used in practice. However the majority of visual media are in 2D format, and most mobile devices are only equipped with 2D camera. Therefore, it is very useful to estimate human pose from 2D image. 

2D pose estimation from images is more difficult than estimation from depth maps
due to ambiguities of appearance and self-occlusion. In general, human pose estimation approaches can be classified into two types: methods based on part-based graphical models, and methods based on regression.
In the first approach using part-based graphical models, the human body structure is embedded into the connections between nodes of the graphical model, and the pose is estimated by finding the pose configuration that best matches the observation as measured by a score function or distribution~\cite{Felzenszwalb05, Eichner2012IJCV, Sapp2010, YiYang2011, Eichner-pami2012, Johnson2011}. 
One popular graphical model for human pose estimation is the pictorial structure model~\cite{Felzenszwalb05} (PSM), which uses pairwise connections between parts to form a tree.  Exact inference is possible and the solution is guaranteed to be globally optimal~\cite{Felzenszwalb05}, but the inference is still very expensive for real-time applications. 
In general, there are two definitions of parts, namely using joints as parts and using limbs as parts. Using joint points as parts avoids the need to predict the orientation of parts, although appearances around joints are more ambiguous.

For both definitions of parts, the appearance model is critical for learning a good PSM~\cite{DGLG13, Eichner2012IJCV}.
Simple appearance models using linear filters are not capable of capturing the parts' appearances, while complicated features are expensive to evaluate at each sliding window. Several methods have been proposed to alleviate this problem by truncating the pose space \cite{Sapp2010, Eichner2012IJCV}. On the other hand, \cite{Yi2013pami} extends the traditional PSM by allowing each body part to have multiple modes.  
Also, multimodal models, such as mixtures of PSM or hierarchical PSM \cite{Johnson2011, Eichner-pami2012, modec13,Pishchulin:2013},  have been proposed. The computation complexity increases rapidly along with the number of modes.    
   
    In the second approach, pose estimation is viewed as a regression task \cite{tgp:2010}.  
These methods train their model to learn a mapping between feature space and pose space. A good feature that encodes pose information is more critical for these methods. Currently, these approaches can only handle small amounts of training data, since calculating a prediction requires solving an expensive optimization problem.
 
In recent years, deep neural network architectures have achieved success in many computer vision tasks~\cite{Sun2013, Farabet2013, Alex2012}. Convolutional neural networks (CNN) are one of the most popular architectures used in computer vision problems because of their reduced number of  parameters compared to fully connected models and intuitive structure, which allows the network to learn translation invariant features. In addition, convolution is a ``cheap'' operation that can be computed efficiently. However, because of the larger capacity (i.e., more parameters) of a deep neural network, it is hard to train a network that generalizes well with limited data.

In this paper, we propose a heterogeneous multi-task framework for human pose estimation using a deep convolutional neural network.  We frame pose estimation as a regression task, while also defining several accessory tasks to guide the network to learn useful features for pose estimation. 
In particular, these accessory tasks are sliding window detectors for different body-parts.
In our framework, the heterogeneous tasks (regression and detection) are trained simultaneously, and we show that the regression network benefits greatly from the accessory detection
tasks, and converges to much better local minima than the network trained with only regression tasks.
We also empirically show that the activation patterns of neurons in the middle layers preserve location information and are selective to localized body-part shapes.

%-------------------------------------------------------------------------

\begin{figure*}[!t]
\begin{center}    
   \includegraphics[width=\linewidth]{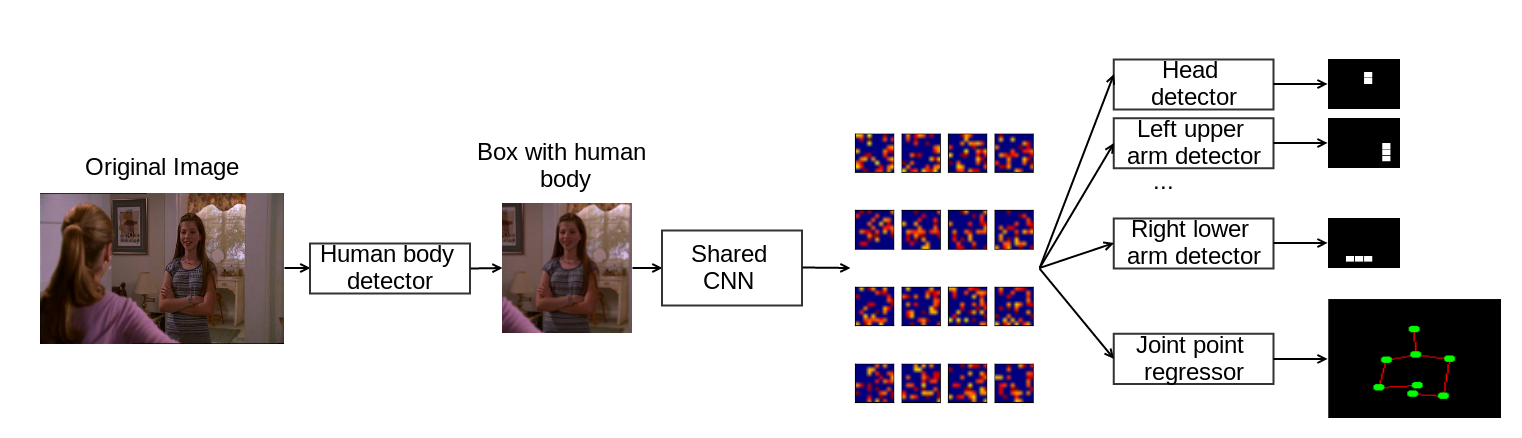}
\end{center}
\vspace{-0.15in}
   \caption{Heterogeneous multi-task learning for pose estimation.
   Given an image, a human body detector is used to find the bounding box around the human.  Next, a convolutional neural network (CNN) extracts shared features from the cropped image, and the shared features are the inputs to the joint point regression tasks and the body-part detection tasks.  
   The CNN, regression, and detection tasks are learned simultaneously, resulting in a shared feature representation that is good for all tasks.
   }
\vspace{-0.2in}
\label{fig:sysdiagram}
\end{figure*}
 
\section{Related work}
Multi-task learning is typically applied when multiple tasks resemble each other and training data for each task is limited \cite{Yu:2005, Yang2009, Evgeniou:2005}. We refer reader to \cite{Yu:2005,Evgeniou:2005} for a review. In the following, we will briefly compare with previous multi-task approaches and regression networks that are most related to our work.
 
In \cite{Yang2009}, a heterogeneous multi-task model is trained by encouraging the parameters for the regression task and the classification task to share the same sparsity pattern. 
They found that joint-training tends to find the most useful features in the input for both tasks. Instead of sharing a sparsity pattern, our framework forces the heterogeneous tasks to share the same feature layers, which results in learning shared feature representation that is good for both tasks.

In \cite{Farabet2013}, a deep convolutional network is trained for scene labeling, by defining a multi-category classification task for each pixel. Instead, we define our detection tasks over sliding windows in the image.
Since we allow each window to contain multiple body parts, each detection task is essentially a binary classification task in a window.

\cite{taylor2010pse} trains a deep CNN to learn a pose-sensitive embedding with nonlinear NCA (neighbourhood components analysis) regression, and predicts the location of the head and hands by finding the nearest neighbor with the learned embedding features. In contrast to \cite{taylor2010pse}, we introduce accessory tasks for learning shared ``pose features'', and output the joint locations directly from the regression network.

In \cite{Sun2013}, a multi-stage system with deep convolutional networks is built for predicting facial point locations. In order to embed a structure prior of the face, they use a set of neural networks that focus on different regions of the input image. 

Similarly, \cite{deeppose2014} trained cascaded convolutional networks for human pose estimation.  Instead of increasing the number of stages for refinement, 
here
 we explore how to improve the performance of a single regression network by introducing accessory tasks. 
Our multi-task strategy could also be used in conjunction with the multi-stage strategy.

In \cite{Weston2008} semi-supervised learning is used to guide the network to learn an internal representation that reflects the similarity between training samples. 
The authors propose that the unsupervised network can either share layers with a supervised network, or serve as an input into the supervised network.
In contrast, we design multiple classification tasks for body parts detection at different location, while all the tasks share the same learned  feature space. 

Finally, in order to investigate the feature representation learned by the neural network, %what have been learned in the neural network, 
\cite{Quoc2012} estimates the ``optimal'' input that maximizes the activation of a selected neuron, and find that the ``optimal'' input resembles a human face.
In contrast to \cite{Quoc2012}, %Instead of looking at a single neuron, 
we visualize a feature by averaging image patches that are associated with the neurons with maximum responses in an upper-layer, and obtain similar results.

\begin{figure}[t]
\begin{center}
   \includegraphics[width=0.9\linewidth]{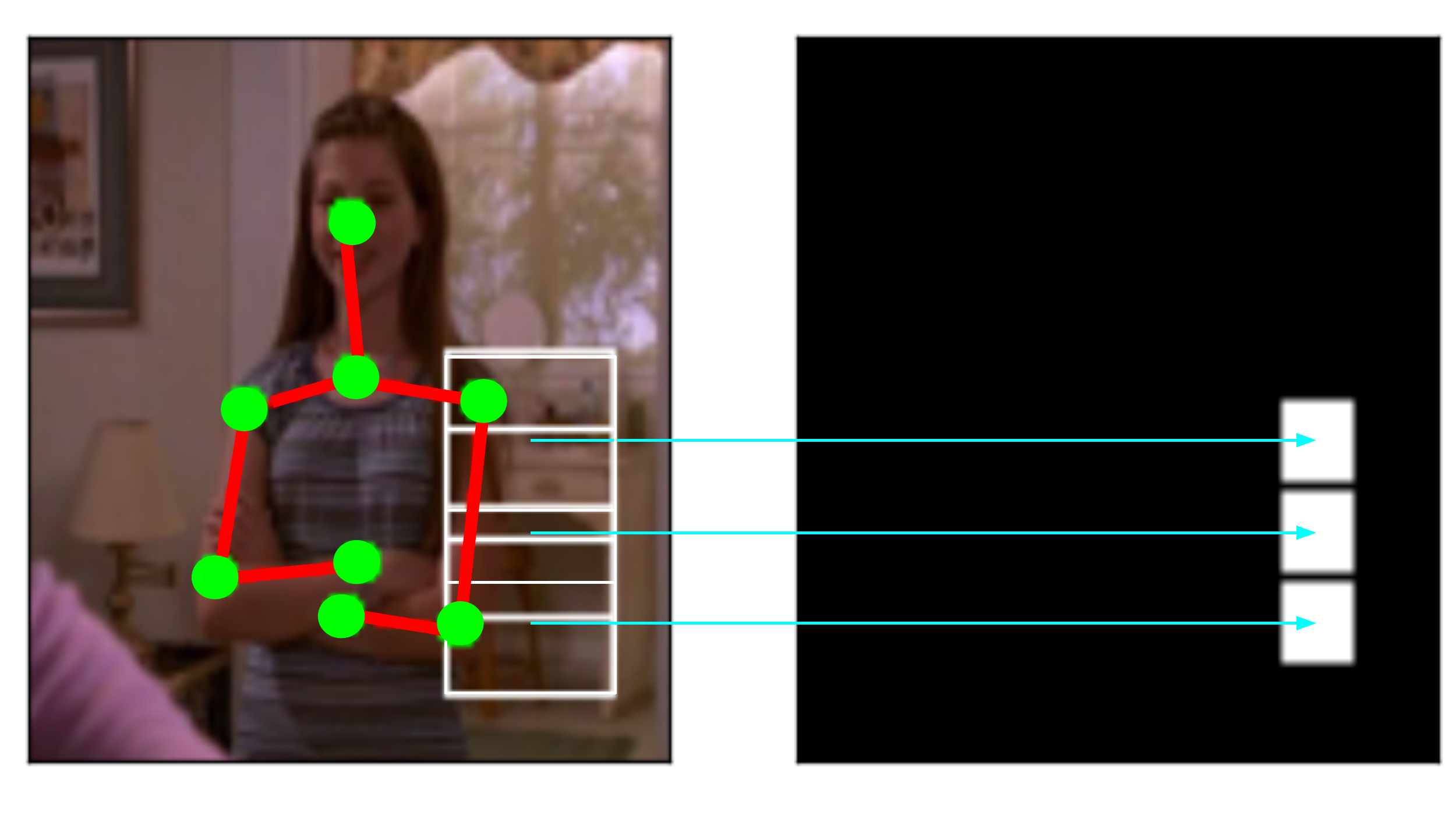}
\end{center}
\vspace{-0.15in}
   \caption{(left) Joint point and body part annotation for pose estimation, and 
(right) the corresponding indicator map for left-upper arm detection.   
In the left image, the green dot indicates the joint point, and the red line is the body part.  The white boxes indicate detection windows that contain the left upper arm.
} 
\label{fig:annotation}
\vspace{-0.2in}
\end{figure}

\section{Heterogeneous Multi-task Learning}
 Our heterogeneous multi-task framework consists of  
 two types of tasks:
 1) a pose regression task, where the aim is to predict the locations of human body joints in an image; 
 2) a set of body-part detection tasks, where the goal is to classify whether a window in the image contains   the specific body part.
 In the following, we assume that a bounding box around the human has already been provided, e.g., using an upper body detector~\cite{ubd}. All the coordinates are with respect to the bounding box containing the human. 
 Our framework is summarized in Figure~\ref{fig:sysdiagram}.

 \subsection{Joint point regression}

 The regression task is to predict the location of %predefined 
 joint points for each human body part. 
 The coordinates of each joint point are taken as the target values. 
 We normalize all the coordinates with the size of bounding box so that their values will be in range of [0, 1].  
 We use the squared-error as the cost function for our regression task,
 \begin{align}
 E_{r}(\hat{J}_{i}, J_i) =  \| J_i - \hat{J}_i \|_{2}^2,
 \end{align}
 where ${J}_i$ and $\hat{J}_i$ are the ground truth and predicted positions for the $i$-th joint, respectively.

 \subsection{Body part detection}

 For the body part detection tasks, the goal is to determine whether a given window in the image contains a specific body part. 
 Let $P$ be the total number of body parts, and let $L$ be the number of overlapping windows inside the bounding box.
 For the $p$-th body part, we train $L$ classifiers, namely $C_{p,1}, ..., C_{p,L}$, to determine whether the $l$-th window contains body part $p$. %, $l \in \{1, \cdots, L\}$. 
 Note that we train a separate classifier for each location $L$, which allows the part detector to learn a location-specific appearance for the part, as well as  location-specific contextual information with other parts.  For example, a lower arm in the upper corner of the bounding box will more likely be vertical or diagonal.

 In our training set, the annotated body parts are represented as sticks.
 Hence, to train the body-part detectors, we need to first identify the windows in the training set that contain each body part.
 A window is considered to contain a body part if the portion of the body part inside the window is at least a particular length, relative to the total length of the part.
 Specifically, we use the following formula to convert the stick annotation of body part $p$ into a binary label indicating its presence/absence in the $l$-th window,
 \begin{align} 
 y_{p,l} = \begin{cases}
 1 ,\quad \text{if} \ \mathrm{len}(window_{l} \cap stick_{p}) > \beta \cdot \mathrm{len}(stick_{p})  \\
 0 , \quad \text{otherwise},
 \end{cases}
 \label{equ:indmap}
 \end{align}
 where $stick_{p}$ is the segment of the $p$-th body part, and $window_l \cap stick_p$ is the portion of $stick_p$ inside $window_l$.  $\beta$ is a fixed threshold, which we empirically set $\beta=0.3$ in all of our experiments. 
 Finally, calculating the binary indicator $y_{p,l}$ for each window $l$, results in a binary indicator map for part $p$. 
 Figure \ref{fig:annotation} shows an example converting the upper-arm annotation into an indicator map.
 Note that  we allow multiple body parts to appear in the same window, and also allow one body part to appear in several windows.   

 For each detection task for part $p$ and window $l$, we minimize the cross-entropy error function,
 \begin{align}
 E_d(\hat{y}_{p, l}, y_{p,l}) = - y_{p, l} \log (  \hat{y}_{p,l}) - (1 - y_{p, l}) \log (1-  \hat{y}_{p,l}), 
 \label{equ:cost-reg}
 \end{align} 
 where $y_{p, l}$ is the ground-truth label, and $\hat{y}_{p, l}$ is the corresponding detection probability from the classifier.

 \begin{figure*}[t]
 \begin{center}    
 \includegraphics[width=0.9\linewidth]{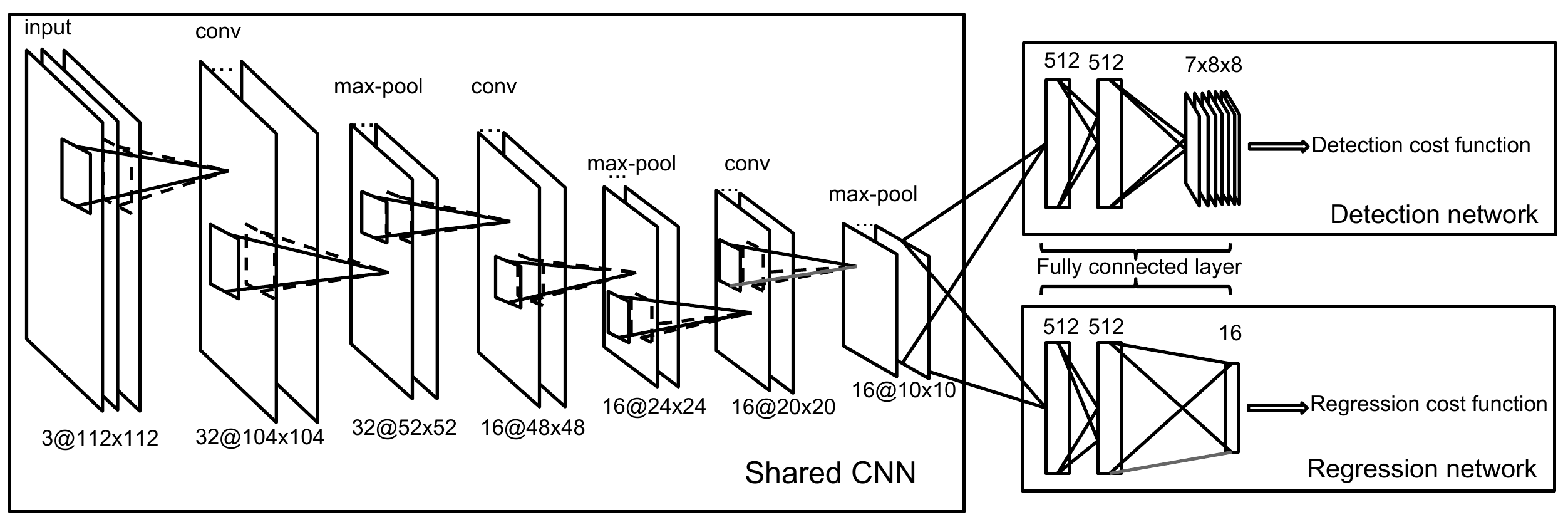} 
 \end{center}
 \vspace{-0.1in}
 \caption{Network architecture for pose estimation: The input layer is $112 \times 112$ RGB image.
 The shared CNN consists of 3 convolutional layers, each followed by a max-pooling layer.  %
 The final pooling layer is connected to separate sub-networks for the joint point regression and body part detection tasks.
 Each subnetwork contains three fully connected layers.
 }
 \vspace{-0.2in}
 \label{fig:network-structure}
 \end{figure*}

 \subsection{Global cost function}
 Our global cost function is the linear combination of the regression cost function for all joints and the detection cost function for all parts and windows, over all training images,
 \begin{align}
 \label{equ:globalcost}
 \Phi = \lambda_{r} \sum_{t} \sum_{i}E_{r}(\hat{J}_{i}^{(t)}, J_i^{(t)})   +
 \lambda_{d}\sum_{t}\sum_{p}\sum_{l} E_{d}(\hat{y}_{p,l}^{(t)}, y_{p,l}^{(t)}),
 \end{align}
 where $\lambda_{r}$ and $\lambda_{d}$ are the weights for regression and detection tasks, respectively, and the superscript $(t)$ indicates the index of the training image.

 \subsection{Network Structure}

 The design of  our network is based on the following considerations:
 \begin{itemize}

 \item {\bf Low level feature sharing}: We allow the detection tasks and regression tasks to share the same learned feature representation.  This is motivated by the following two reasons.
 First, features learned for the detection task should also be helpful for identifying parts or joints in the regression task. Second, feature sharing will reduce the number of parameters and encourage the network to generalize on a larger range of samples. 
\vspace{-0.2in}
 \item {\bf Preservation of location information}:  The detection task is to determine whether a local window contains the specific body part, while the regression task is to predict the coordinates of the joint position.  Hence, the features extracted from the lower layers should not be translation invariant, i.e., the positions of the features should be preserved in the feature map. 
 \vspace{-0.1in}

 \item {\bf Integration of context information}: Sometimes it is difficult to distinguish different body parts by only looking at the bounding box of the body parts. 
 For example, when wearing long-sleeves, the upper arm and lower arm can have very similar appearance, and hence it is hard to distinguish them by only looking at the windows containing these two parts.  Including context information about neighboring parts can help to improve the part detector.  Hence, the input for each local part detector is the whole bounding box image (the whole human).
 \vspace{-0.1in}

 \end{itemize}

 Our network structure is shown in Figure~\ref{fig:network-structure}.  The input is an RGB image with human.
 The first 6 hidden layers are shared by both regression and detection tasks. 
 In the shared layers, we only use convolutional layers and pooling layers to ensure the activation of neurons are affected by only local patterns in the input. 
 We also choose to use a small filter and stride size to keep more location information. 

 Each convolutional layer consists of several maps. Filter weights are shared within each map, which means the neurons within the same map are sensitive to %``expecting'' 
 the same patterns at different location in the previous layer.   Neurons at the same position (but belonging to different maps) will always contribute to the same unit in the next layer.
 The max-pooling layer is added after each convolutional layer to increase non-linearity and to integrate local information. 

 The value of neuron $i$ in a convolutional layer or regression layer is calculated by
 \begin{equation}
\vspace{-0.11in} 
 v(i) = f_{act}( \sum_{j \in R_{i}}w_{i,j} v(j)  ),
 \end{equation}
 where $R_{i}$ is the set of neurons from which neuron $i$ receives input, $w_{i,j}$ is the weight between neuron $i$ and neuron $j$, and $f_{act}$ is the activation function of that layer.  
 Most of the neurons in our network are Rectified Linear Units (ReLu)~\cite{relu2010}, where $f_{act}(x) = \max(0,x)$. \cite{relu2010} showed that ReLus are good for recognition tasks and fast to train. 
 We use the hyperbolic tangent as the activation function in the last layer of the regression task, and the logistic function in all the last layer of detection tasks.

\vspace{-0.06in}
 \subsection{Training}
 \vspace{-0.08in}
We jointly train the regression and detection networks with the global cost function in (\ref{equ:globalcost}).
 We use back-propagation~\cite{Lecun98} to update the weights. 
Given a training image, predictions for both tasks are calculated, and the corresponding gradients are back-propagated through the network.
 For layers with several output layers, the gradient from their output layers are summed together for weight updating. ``Dropout'' \cite{dropout2012} is also used in the first fully connected layers for the regression and detection tasks to prevent over-fitting. The dropout probability is set to be 0.5 in the experiments.  In each iteration, the neurons in dropout layers will be randomly selected with probability 0.5 to forward their activation to the output units, and only the selected neurons will participate in the back-propagation during this iteration. In the testing stage, all the neurons are used for prediction with their activation value multiplied by 0.5 for normalization. This strategy turns out to be very effective, since without ``dropout'', our network will severely overfit. We refer reader to \cite{Alex2012} for more details about the training procedure.

 \section{Experiments}
 \vspace{-0.1in}
 We present  experiments using our  method HMLPE (heterogeneous multi-task learning for pose estimation).

 \vspace{-0.05in}
 \subsection{Training data}
 We collect training data from several data sets, including Buffy Stickmen~\cite{Eichner2012IJCV}, ETHZ Stickmen~\cite{Eichner2009BMVC}, Leed Sport Pose (LSP~\cite{Johnson2011}), Synchronic Activities Stickmen (SA~\cite{Eichner-pami2012}),  Frames Labeled In Cinema (FLIC~\cite{modec13}),  We Are Family(WAF)~\cite{eichner2010}. 
 For Buffy, LSP, FLIC we only use their respective training sets, while 
 we use the whole ETHZ, SA, and WAF datasets for training.
 In total, we have collected 8427 images for training. 

 We represent the human body with a set of joints, and use the segments between those joints to represent body parts. For data sets with only stick labels, we use the nearest end of stick or average of nearest ends as the joint point.
 We define 8 joints (nose, neck, left and right shoulders, left and right elbows, and left and right wrists), and 7 body parts (head, left and right shoulder, left and right upper arms, and left and right lower arms). 
 Since Buffy, ETHZ, SA, WAF only provide the upper-end and lower-end of the head, we use the middle point as the nose position. 
 We illustrate our parts and joints definition in Figure~\ref{fig:annotation}.

 Bounding boxes for the training images are generated according to the ground-truth labels.  
 We select a bounding box for each training image that contains all the annotated body parts, 
 and then resize the image inside the bounding box to $128 \times 128$.
 We then augment the dataset by randomly selecting 16 bounding box of size $112 \times  112$ inside the extracted human image, and apply a mirror transformation to double the training set. In total, the training set is augmented by a factor of 32. 

 In the current experiments, images with occluded body parts are removed, 
 although our framework could be extended to handle training poses with occlusion. %}
 \vspace{-0.02in}
 \subsection{Experiment setup}

 For our HMLPE, the pose regression task predicts 8 joint positions (16 outputs in total), and the detection task has 7 body parts.
 For the detection task in HMLPE, we use 64 local windows of equal size %(except windows near the boundary)
 %with their center being 
 uniformly distributed in the bounding box.  The window size is set to $30 \times 30$ in all experiments,
 which is % which is determined empirically to make it 
 comparable to the size of a body parts found in the training set.
 We pre-train the network using the training data discussed in the previous section, in order to obtain an initial network.
 Then, we use the initial network as the starting point for training the network using the training data of a specific dataset, either Buffy or FLIC.
 The initial network serves as a prior to help regularize the network.
 We train and evaluate our network on a Dell T3400 with GTX 770 4G. 
 Training the network takes 1 to 2 days, while the evaluation for 4000 images takes 5-6 seconds.
% 

 % 

 %

 % ////////

 % FIG HERE
 % fig:response

 % ////////

 % ///////////
 % FIG HERE
 % detector

 % 
 % 
 % 
 % 

 \vspace{-0.02in}

 \subsection{Evaluation on Buffy Set}

 \begin{table}
 \small
 \begin{center}
 \caption{PCP on Buffy test set.  LL, RL, LU, and RU mean left-lower, right-lower, left-upper, and right-upper.}
 \label{tab:pcp-buffy}
 \begin{tabular}{|c|c|c|c|c|}
 \multicolumn{5}{c}{whole test set (276 images)}
 \\
 \hline
 PCP ($\alpha = 0.5$) & LL arms  & RL arms& LU arms & RU arms \\
 \hline
 %HMLPE (pre-train) & 55.43 & 54.71 & 89.86 & 93.12 \\ \hline 
 HMLPE  & 55.80 & 56.88 & 90.22 & 93.12 \\ \hline
 RoDG-Boost\cite{Hara2013} & \multicolumn{2}{c|}{51.5} & \multicolumn{2}{c|}{92.8} \\ \hline
 Eichner\cite{Eichner2012IJCV} &  \multicolumn{2}{c|}{50.0} & \multicolumn{2}{c|}{81.9} \\ \hline
 MoP \cite{Yi2013pami} $M=6$  & 51.45 & 55.43 & 82.25 & 87.68 \\ \hline 
 MoP \cite{Yi2013pami} $M=9$  & 56.52 & 55.80 & 84.78 & 89.13 \\ \hline 
 MoP \cite{Yi2013pami} $M=12$ & 60.87 & 59.78 & 85.87 & 88.41 \\ \hline 
 \multicolumn{5}{c}{test subset with correct upper-body detections (267 images)}\\ \hline
 HMLPE  & 57.68 & 58.80 & 93.26 & 96.26 \\ \hline
 MoP \cite{Yi2013pami} & \multicolumn{2}{c|}{57.5} & \multicolumn{2}{c|}{94.3}\\ \hline 
 \end{tabular}
 \vspace{-0.3in}
 \end{center}
 \end{table}

 We use the same upper body detector as \cite{Eichner2012IJCV, Hara2013}. In order to get the human bounding box, the width and height of the upper body detection windows are scaled by a fixed factor ($s_{width}=1.7$, $s_{height}=4.2$), which were empirically set according to the training set.
 %.
 The scaled detection window is used as the human bounding box, and the image is cropped and resized to $112 \times 112$. 

 We use Percentage of Correct Part (PCP) to measure the accuracy of pose estimation. As pointed out in \cite{Hara2013}, the previous PCP evaluation measure does not compute PCP correctly. 
 We use the evaluation tool provided by \cite{Hara2013} to calculate the corrected PCP,
 %.
 where an estimated body part with end points $(e_{1}, e_{2})$ is considered as correct if
 \begin{equation}
 \begin{split}
 \| e_{1} - g_{1} \|_{2} \le \alpha \cdot L & \ \  \mathrm{and}\ \  \| e_{2} - g_{2} \|_{2} \le \alpha \cdot L \\
 & \mathrm{or} \\
 \| e_{2} - g_{1} \|_{2} \le \alpha \cdot L & \ \  \mathrm{and}\ \  \| e_{1} - g_{2} \|_{2} \le \alpha \cdot L 
 \end{split}
 \end{equation}
 where $(g_{1}, g_{2})$ and $L$ are ground truth position and length of the part, and $\alpha$ is the parameter for PCP.  We use the standard value of $\alpha=0.5$.

 Table~\ref{tab:pcp-buffy} presents the PCP results of lower and upper arms (since we have different definitions of torso and head parts, we do not show the evaluation here).
 On the whole Buffy test set (276 images), HMLPE achieves better results than \cite{Hara2013, Eichner2012IJCV} on the more difficult parts, lower arms (4.8\% improvement), but gets a slightly worse result than \cite{Hara2013} on upper arms (1.1\% lower) .
 Evaluation on the whole Buffy test set includes errors due to mis-detection of the upper body.
 To investigate the pose estimation performance alone, we also present results on the subset of the Buffy test set where the upper body detector predicts the correct bounding box. In this case, HMLPE achieves slightly better results than \cite{Yi2013pami} (0.7\% better on lower arms and 0.5\% better on upper arms).
 We also run the code from \cite{Yi2013pami} on our full training set, 
 using different number of components per part (denoted as $M=\{6,9,12\}$).
 The default setting of $M=6$ gets worse results than the model trained with only the Buffy training set, 
 most likely because the full training set contains more variance in poses.
 Increasing the number of components improves the accuracy, but at an increased cost of training, e.g., 4 days were needed to train the $M=9$ model.
 Using $M=12$, \cite{Yi2013pami} has better PCP (4\%) on the lower arms compared to HMLPE.  On the other hand, HMLPE has better PCP (4.5\%) than \cite{Yi2013pami} on upper arms. 
 %
 %
 %
 %

 %

 %
 %
%%%%%
  
\subsection{Evaluation on FLIC Data set}
 
Next we evaluate on the FLIC test set.
We use the same torso box as \cite{modec13} with scale 
%ght and weight of torso box by a fixed 
factors ($s_{width} = 3.5$, $s_{height} = 4.5$) set empirically from the training set.
%,
\cite{modec13} uses the following accuracy to evaluate their performance,
\begin{equation} 
acc_{{J}_{i}}(r) = \frac{100}{N_{sample}} \sum_{t = 1}^{N_{sample}} \pmb{1}\left(\frac{100\cdot \| {J}_{i}^{(t)} - \hat{J}_{i}^{(t)} \|_{2}}{\|{J}_{lhip}^{(t)} - {J}_{rsho}^{(t)}   \|_{2}} \le r\right).
\end{equation} 
where ${J}_{i}^{(t)}$ and $\hat{J}_{i}^{(t)}$ are the ground truth annotation and predicted position for the $i$-th joint point of test image $t$.

\begin{figure}
\begin{center} 
\includegraphics[width=0.9\linewidth,height=0.9\linewidth]{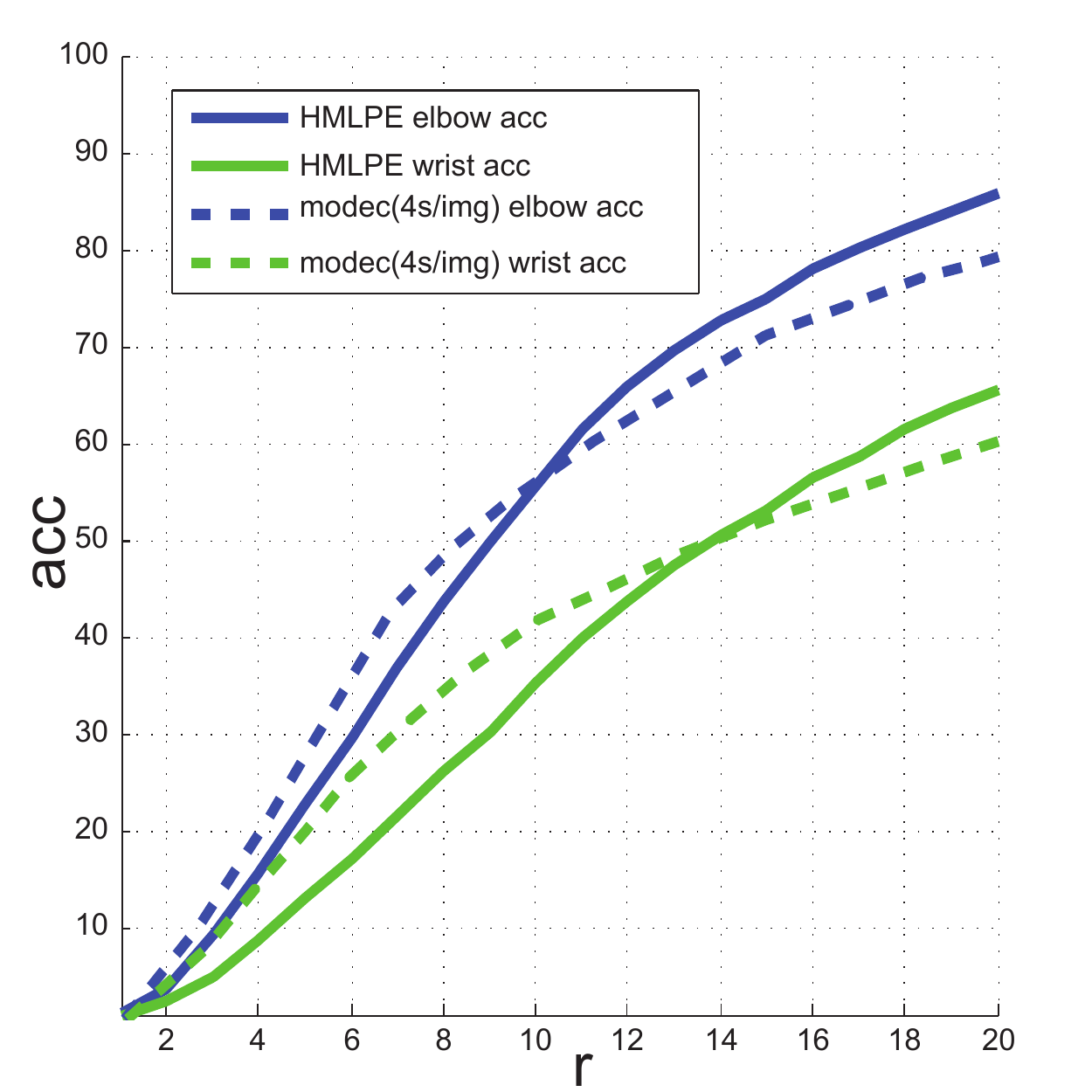}
%% HERE FONT
\end{center} 
\vspace{-0.1in}
   \caption{Test results on the FLIC data set.}
\vspace{-0.10in} 
\label{fig:flic-compare} 
\end{figure}

Since \cite{modec13} compares their methods with several previous approaches, and show that their model performs the best under this criteria, we only compare with \cite{modec13}. 
The accuracy results are shown in Figure~\ref{fig:flic-compare}. 
HMLPE has better accuracy with a looser criteria (larger $r$); for $r=20$, the accuracy of HLMPE on wrists and elbows is about $6\%$ higher than MODEC.
On the other hand, HLMPE has worse accuracy with a strict criteria (when $r=6$, HMLPE is about $7\%$ and $5\%$ lower than MODEC on wrists and elbows). These results suggest that HLMPE can robustly estimate the general pose, but is less accurate at estimating the exact location of each joint. 
Also, we have trained MODEC on our full training set, but did not observe any improvement.

In addition, we measure PCP on the FLIC dataset to facilitate future comparisons (see Table~\ref{tab:pcp-flic}).

\begin{table}
\begin{center}
\small
\caption{PCP on FLIC test set.}
\label{tab:pcp-flic}
\begin{tabular}{|c|c|c|c|c|}
\hline
PCP( $\alpha = 0.5$) & LL arm  & RL arm& LU arm & RU arm \\
\hline 
HMLPE &  59.05 & 56.70 &92.81 & 92.52 \\ \hline
\end{tabular}
\vspace{-0.3in} 
\end{center}
\end{table}

\subsection{Effect of multi-task training}
\begin{figure*}[t]
 \begin{center}  \small
 \begin{tabular}{cccc}
 \multicolumn{2}{c}{\quad \quad pose regression task} &  \multicolumn{2}{c}{part detection tasks} \\
 \parbox{0.24\textwidth}{ \quad  \qquad \qquad   training error} & \parbox{0.2\textwidth}{\qquad \quad test error} & \parbox{0.2\textwidth}{\quad\qquad training error} & \parbox{0.25\textwidth}{\quad\qquad test error} \\
 \multicolumn{4}{l}{\includegraphics[width=0.9\linewidth, height=0.4\linewidth]{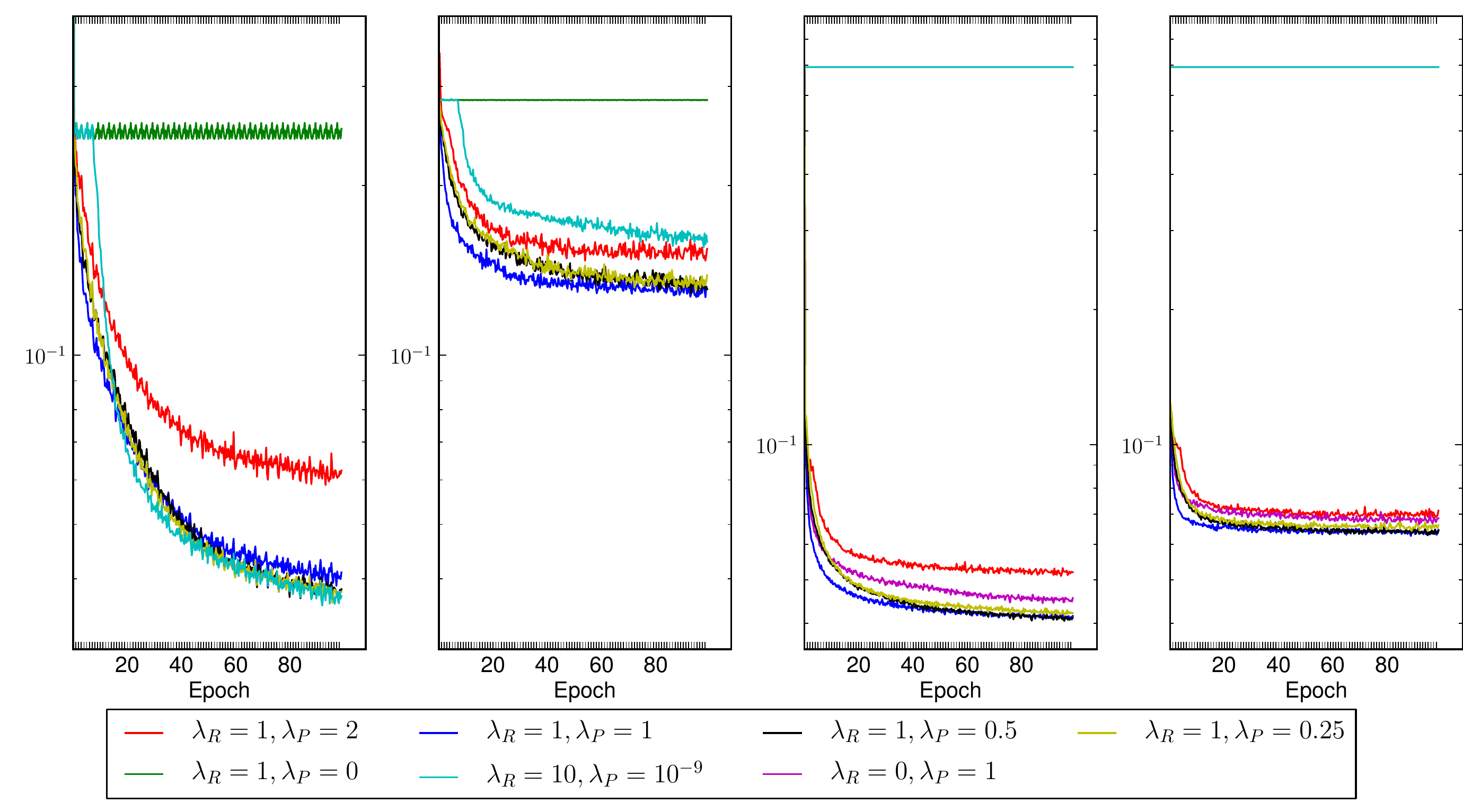}}
 %
 %  \\
 %
 %
 %
 \end{tabular}      
 \caption{Effect of changing the weights in multi-task learning: training and test errors for (left) the pose regression task, and  (right) the detection tasks. The test errors are the average costs of regression (left) and detection (right) tasks on the Buffy and FLIC test datasets.}
 \vspace{-0.3in}
 \label{fig:net-structure-comparision}
 \end{center}      
 \end{figure*}

%%%%% 
Next we study the effect of multi-task training, i.e., the joint learning of the regression and detection tasks.
We set different values for the weights of the regression and detection tasks. 
All parameters except the weights on the cost function are kept the same.
We %
show
training and testing error in Figure~\ref{fig:net-structure-comparision} and in Table~\ref{tab:net-structure-comparison}.

Firstly, the network with only the regression task performs poorly on both the training and testing sets.\footnote{Training the network with different initializations gave similar results.}  Even using tiny weights on the detection tasks help to improve 
the convergence, leading to a significant performance increase.
Within a certain range, increasing weights on the detection tasks leads to lower errors on the test set.
%}
 For larger weights on the detection tasks, the performance decreases. 
This is reasonable since the gradient will be dominated by detection task in this case.

These results suggest that the regression task benefits greatly from the feature representation induced by the detection tasks. 
The gradients from detection tasks not only guide the network to converge to a better minimum on the training set, but also help to enhance the generalization.
Although the network needs to learn 7*8*8 detectors from limited training data, sharing features among the detection and regression tasks seems to be an effective way for learning useful features for both tasks.

\begin{table}[tbhp]
% \vspace{-0.15in}
\caption{Effect of changing the weights on each task - the training and testing errors are for epoch 100.}
\tabcolsep=0.07cm 
\small
\begin{tabular}{|c|c|c|c|c|c|c|c|}
\hline
$\lambda_{R}/\lambda_{P}$&0  &0.5 & 1   & 2   & 4   &  $10^{10}$& $\infty$ \\
\hline
 training error (R)&  -    &0.059&0.039&0.036&0.036&0.036     &0.241\\ 
 test error (R)    &  -    &0.149&0.131&0.132&0.137&0.163     &0.284\\
\hline
 training error (P)& 0.045 &0.051&0.041&0.041&0.042&0.693     & -   \\
 test error (P)    & 0.069 &0.070&0.064&0.064&0.066&0.693     &     -   \\
\hline
\end{tabular}
\label{tab:net-structure-comparison}
\end{table}

\vspace{-0.1in}
\section{Visualization of features}
\label{sec:visualization}

\begin{figure*}[t]
\begin{center}
\begin{tabular}{cc}
 \multicolumn{2}{c}{ \raisebox{0.35in}{(a)}  \includegraphics[width=0.8\linewidth]{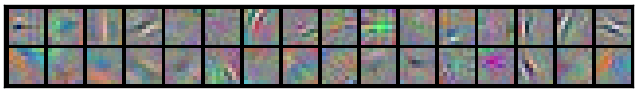} }\\
\raisebox{0.45in}{(b)} \includegraphics[width=0.43\linewidth]{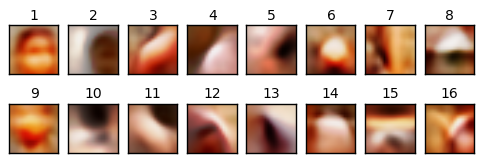} &
\raisebox{0.45in}{(c)}    \includegraphics[width=0.43\linewidth]{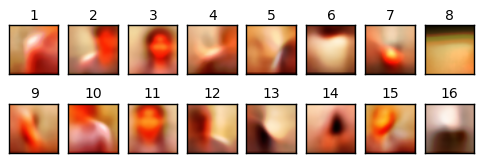} \\
\end{tabular}

\caption{Visualization of low-, mid-, and high-level features in our trained network: (a) shows 32 filter weights in the first convolutional layer; visualizations of the (b) mid-level features from the second convolutional layer; and (c) high-level features from the third convolutional layer. 
}
\vspace{-0.3in}
\label{fig:vt}
\end{center}

\begin{center}

\begin{tabular}{cc}

\includegraphics[width=0.5\linewidth]{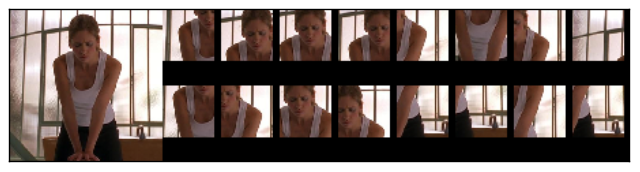} & 
\includegraphics[width=0.5\linewidth]{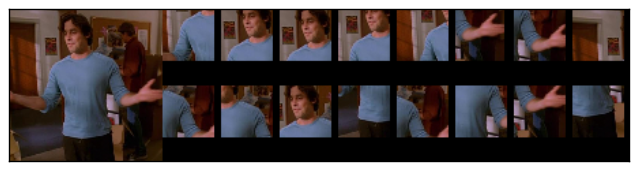} \\
\end{tabular}
  \caption{Examples of backtracked patches in the original image. Each image patch is the backtracked  patch that caused maximum activation in a feature map of the 3rd convolutional layer. The order of patches corresponds to the order of features in Fig.~\ref{fig:vt}c.}
\label{fig:backtrackingshow}
\vspace{-0.4in}
\end{center}

\end{figure*}

In this section, we investigate the features learned by the network. 
Since the first convolutional layer operates on the input image, the filter response can reflect what low-level patterns in the image to which the neurons are sensitive. The learned filters are in Fig.~\ref{fig:vt}a, and as expected, they look like edge or gradient detectors for different orientations.

For the 2nd and 3rd layers (mid- and high-level features), we use a different approach than 
\cite{Quoc2012}, which finds the input that maximizes one specific neuron.
Instead, we use the property that our network is only locally connected in the first 6 layers.
That is, the activation of some neurons in the middle layers are only affected by a sub-region of the input image. 
In addition, the connection is regular, we can backtrack through the network to find the region of the image from which a neuron received its input. 
We present the backtracking algorithm in Algorithm~\ref{algo:backtrack}. 
Since filter weights are shared within the same feature map, neurons in the same map are ``expecting'' the same local patterns in the previous layer. 
Based on these properties, we consider the activation of one feature map at a time. Instead of solving an optimization problem, we select the patches in the original image that contribute to the maximum activation in one feature map. 
Figure~\ref{fig:backtrackingshow} shows the backtracked patches on a Buffy test image for different features in the 3rd convolutional layer.
Surprisingly, we find some feature maps work like body part detectors --- the maximal activation in some maps occurs more frequently on neurons that take inputs from region of body parts, such as head, shoulders and arms.

To visualize the feature of a map, we average all its corresponding backtracked patches from all training images.
The average backtracked patches for each map in the 2nd and 3rd convolutional layers are shown in Figure~\ref{fig:vt}b and \ref{fig:vt}c.
 The average backtracked patches show more clear patterns of body parts like head, shoulder, upper arms. In particular, the visualizations of the mid-level features in Fig.~\ref{fig:vt}b look like body part detectors, such as head (feature 1), neck (feature 9), arms (feature 5), and shoulders (feature 14).
Similarly, the high-level features in Fig.~\ref{fig:vt}c look like {\em localized} body parts, e.g., heads in different positions (features 2, 3, and 11), left and right shoulders (features 1 and 10), and arms (features 6, 9, and 15).  There are also a few high-level features that do not correspond to specific body parts.  For example, feature 8 in Fig.~\ref{fig:vt}c has two horizontal bands of color, and appears to respond to horizontal background structures, such as windows and the tops of door frames (see Fig.~\ref{fig:backtrackingshow}).  This feature could be useful for identifying context information, such as the location of the top of the door relative to the top of the head.

\begin{algorithm}[b!]

\caption{Backtracked patches}
\small
\label{algo:backtrack}
\begin{algorithmic}
\REQUIRE $layer\_list, R = (mx,my,mx,my) $

\COMMENT{(mx,my) are the location of maximum activation}
\FOR{$l$ in reversed($layer\_list$)}
\STATE $R_{lx} \gets R_{lx} \cdot l.stride$
\STATE $R_{ly} \gets R_{ly} \cdot l.stride$
\STATE $R_{ux} \gets R_{ux} \cdot l.stride + l.filter\_size - 1$
\STATE $R_{uy} \gets R_{uy} \cdot l.stride + l.filter\_size -1$
\ENDFOR
\end{algorithmic}
\end{algorithm}

% ///////////

\vspace{-0.1in}
\section{Conclusion}
\vspace{-0.05in}
In this paper, we have proposed a heterogeneous multi-task learning framework with deep convolutional neural network for human pose estimation. Our framework consists of two tasks: pose regression and body-part detection via sliding-window classifiers.
We empirically show that jointly training pose regression with the detection tasks  guides the network to learn meaningful features for pose estimation, and makes the network generalize well on testing data.
Finally, we visualize the mid- and high-level features using the average of backtracked patches from the maximally responding neurons.  We found that these neurons are selective to shape patterns resembling localized human body parts.

In  future work, we will extend our network for learning poses with occlusion, and combine our framework with unsupervised learning for pre-training the network.
In addition, we would like to extend our framework for estimating human pose from video sequences, as well as other structured objects.

\begin{acknowledgements}
\vspace{-0.1in} 
This work was supported by the Research Grants Council of the Hong Kong Special Administrative Region, China (CityU 123212, CityU 118810, and CityU 119313).
\vspace{-0.1in}
\end{acknowledgements}

\vspace{-0.1in}
{\small
\bibliographystyle{ieee}
\bibliography{refs}
}

\end{document}